\documentclass{article}
\usepackage{arxiv}

\usepackage{amsthm}

\usepackage{xcolor}  

\newcommand{\todo}[1]{{\color{red} #1}}

\usepackage{amssymb}

\usepackage{amsmath}
\usepackage{mathdots}

\usepackage{mathtools}

\usepackage{tensor}

\usepackage{urwchancal}
\DeclareFontFamily{OT1}{pzc}{}
\DeclareFontShape{OT1}{pzc}{m}{it}{<-> s * [1.10] pzcmi7t}{}
\DeclareMathAlphabet{\mathpzc}{OT1}{pzc}{m}{it}

\usepackage{graphicx}
\usepackage{ifpdf}
\ifpdf
\usepackage{epstopdf}
\PrependGraphicsExtensions{.svg}
\fi

\newtheorem{theorem}{Theorem}[section]

\newtheorem{proposition}[theorem]{Proposition}

\newtheorem{definition}[theorem]{Definition}

\providecommand{\R}{\mathbb{R}}

\providecommand{\SO}{\mathbf{SO}}

\providecommand{\SE}{\mathbf{SE}}

\providecommand{\grpG}{\mathbf{G}}

\providecommand{\gothso}{\mathfrak{so}}

\providecommand{\gothg}{\mathfrak{g}}

\providecommand{\gothX}{\mathfrak{X}} 

\providecommand{\so}{\mathfrak{so}}
\providecommand{\se}{\mathfrak{se}}

\providecommand{\calG}{\mathcal{G}}

\providecommand{\calM}{\mathcal{M}}
\providecommand{\calN}{\mathcal{N}}

\providecommand{\vecL}{\mathbb{L}}

\providecommand{\tT}{\mathrm{T}} 

\providecommand{\Id}{I} 

\DeclareMathOperator{\Ad}{Ad}

\providecommand{\tD}{\mathrm{D}}

\providecommand{\mr}[1]{\mathring{#1}} 

\usepackage{accents}
\makeatletter
\providecommand{\scirc}{%
    \hbox{\fontfamily{\rmdefault}\fontsize{0.4\dimexpr(\f@size pt)}{0}\selectfont{\raisebox{-0.52ex}[0ex][-0.52ex]{$\circ$}}}}

\makeatother

\mathchardef\mhyphen="2D

\usepackage{graphicx}      
\usepackage{natbib}        
\usepackage{hyperref}
\usepackage{algorithm,algpseudocode}

\newcommand{\papercitation}{
Hampsey, M., van Goor, P., Hamel, T., Mahony, R. (2022), ``Exploiting Different Symmetries for Trajectory Tracking Control with Application to Quadrotors'', \emph{Accepted for presentation at IFAC NOLCOS}.
}

\newcommand{\publicationdetails}
{
\copyrightNoticeIFACAccepted{2022} 
This research was supported by the Australian Research Council
through Discovery Grant DP210102607 `` Exploiting the Symmetry of
Spatial Awareness for 21st Century Automation'' and the Franco-Australian International Research Project “Advancing Autonomy for Unmanned Robotic Systems” (IRP ARS).
\papercitation 
}
\newcommand{\publicationversion}
{Author accepted version}

\providecommand{\todo}[1]{\todo{#1}}
\begin{document}

\title{Exploiting Different Symmetries for Trajectory Tracking Control with Application to Quadrotors}
\headertitle{Exploiting Different Symmetries for Trajectory Tracking Control}

\author{
\href{https://orcid.org/0000-0001-2345-6789}{\includegraphics[scale=0.06]{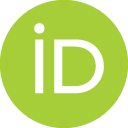}\hspace{1mm}
Matthew Hampsey}
\\
	Systems Theory and Robotics Group \\
	Australian National University \\
    ACT, 2601, Australia \\
	\texttt{matthew.hampsey@anu.edu.au} \\
	\And	\href{https://orcid.org/0000-0001-2345-6789}{\includegraphics[scale=0.06]{orcid.png}\hspace{1mm}
Pieter van Goor}
\\
	Systems Theory and Robotics Group \\
	Australian National University \\
    ACT, 2601, Australia \\
	\texttt{pieter.vangoor@anu.edu.au} \\
\And	\href{https://orcid.org/0000-0002-7779-1264}{\includegraphics[scale=0.06]{orcid.png}\hspace{1mm}
    Tarek Hamel}
\\
    I3S (University C\^ote d'Azur, CNRS, Sophia Antipolis) \\
    and Insitut Universitaire de France \\
    \texttt{THamel@i3s.unice.fr} \\

\And	\href{https://orcid.org/0000-0001-2345-6789}{\includegraphics[scale=0.06]{orcid.png}\hspace{1mm}
Robert Mahony}
\\
	Systems Theory and Robotics Group \\
	Australian National University \\
	ACT, 2601, Australia \\
	\texttt{robert.mahony@anu.edu.au} \\
}

\maketitle

\title{Exploiting Different Symmetries for Trajectory Tracking Control with Application to Quadrotors\thanksref{footnoteinfo}}

\begin{abstract}                
High performance trajectory tracking control of quadrotor vehicles is an important challenge in aerial robotics.
Symmetry is a fundamental property of physical systems and offers the potential to provide a tool to design high-performance control algorithms.
We propose a design methodology that takes any given symmetry, linearises the associated error in a single set of coordinates, and uses LQR design to obtain a high performance control; an approach we term Equivariant Regulator design.
We show that quadrotor vehicles admit several different symmetries: the direct product symmetry, the extended pose symmetry and the pose and velocity symmetry, and show that each symmetry can be used to define a global error.
We compare the linearised systems via simulation and find that the extended pose and pose and velocity symmetries outperform the direct product symmetry in the presence of large disturbances. This suggests that choices of equivariant and group affine symmetries have improved linearisation error.
\end{abstract}



\section{Introduction}

Autonomous drones have steadily grown in popularity over past decades, with modern applications in a diverse array of domains, including photography, cinematography, surveying, inspection, delivery, and racing.
Quadrotors are a particularly compelling research and development platform owing to mechanical simplicity and versatile flight dynamics.
A key goal in quadrotor control is tracking a desired trajectory in the presence of external disturbances and state perturbations.
This problem has a long history in general robotics and several classical control design methodologies are available.
One of the most studied approaches is the application of the Linear Quadratic Regulator (LQR) in trajectory error coordinates (\cite[chapter 4. Tracking Systems]{anderson2007optimal}).
However, classical LQR is posed for systems on $\R^m$ whereas the natural state space of a quadrotor is a smooth manifold.

If a dynamical system is posed on a smooth manifold that admits a Lie group symmetry, then the group symmetry can be used to define error coordinates for control design.
This approach is powerful, and was originally used to derive an almost-globally stable non-linear attitude controller on $\SO(3)$ (\cite{Meyer1971}) and later a quaternion-based attitude controller (\cite{Wie1985}).
This approach was also more recently applied to aerial robotics by utilising a symmetry on $\SO(3) \times \R^3 \times \R^3$ to design a nonlinear cascaded controller to track attitude and position (\cite{2010_Lee_Geometric}).

Using a Lie group structure also allows one to work in local logarithmic error coordinates in a single chart.
In \cite{2019_Farrell_Error}, the authors consider the quadrotor system placed on the direct product Lie group $\SO(3) \times \R^3 \times \R^3$.
The corresponding local logarithmic coordinates are used to track a trajectory with a standard LQR formulation.

In parallel to the control work, significant progress has been made in exploiting symmetry for observer and filter design.
In \cite{mahony2008}, the Lie group structure of $\SO(3)$ is used to design a complimentary attitude filter.
In \cite{bonnabel2008}, an invariant observer is proposed by exploiting the structure of the unit quaternions.
The Lie group $\SE_2(3)$ is introduced in \cite{2014_Bonnabel} to design a filter for inertial navigation. This work is further developed in the landmark paper \cite{2017_Bonnabel}, which proves that the local logarithmic linearisation is exact for \emph{group affine} systems.
More recent work in this field shows that group affine and \emph{equivariant} symmetries lead to simplified error dynamics and improved linearisation error (\cite{Bonnabel_2019}, \cite{Mahony_2022}, \cite{vangoor2021equivariant}).

The choice of symmetry is important, because it determines the non-linear error dynamics and, to a large extent, the quality of the subsequent linearisation.
In recent work, \cite{2020_Forbes_Quadrotor} use the $\SE_2(3)$ symmetry for (local) control of a quadrotor system: this is the first time a symmetry other than $\SO(3)$ (or a direct product of $\SO(3)$) was used for quadrotor control.

In this paper we propose the `Equivariant Regulator' (EqR), a general design methodology for trajectory tracking on symmetry groups.
We show that a trajectory on a manifold equipped with a free and transitive group action can be lifted onto the corresponding Lie group.
We show that if the system is equivariant, we can formulate a global error for which the error dynamics can be written in terms of only the error state and transformed system inputs.
We derive the general form of the linearisation in logarithmic coordinates.
Exploiting the generic design methodology proposed, we consider the same control algorithm implemented for different choices of Lie group symmetry, and compare the resulting tracking performance in simulation.

\section{Preliminaries}
\label{sec:preliminaries}
\vspace{-2mm}
\subsection{Notation}
\vspace{-2mm}
For a thorough introduction to smooth manifolds and Lie group theory, the authors recommend \cite{2012_lee_SmoothManifolds} or \cite{tu2010introduction}.

Let $\calM$ be a smooth manifold.
For arbitrary $\xi \in \calM$, the tangent space of $\calM$ at $\xi$ is denoted by $\tT_{\xi}\calM$.
For another smooth manifold $\calN$ and a smooth mapping $h : \calM \to \calN$,
\begin{align*}
\tD_{\xi | \zeta}h(\xi) : &\tT_{\zeta}\calM \to \tT_{h(\zeta)}\calN\\
&\delta \mapsto \tD_{\xi | \zeta}h(\xi)[\delta]
\end{align*}
denotes the differential of $h(\xi)$ evaluated at $\xi = \zeta$ in the direction $\delta \in \tT_{\zeta}\calM$.
The notation $\tD h$ will also be used when the argument and base point are implied.
The space of smooth vector fields on $\calM$ is denoted with $\gothX(\calM)$.

Let $\grpG$ be a Lie group.
The identity element is denoted $I$.
If $\grpG$ is an $m$-dimensional Lie group,  $\tT_X \grpG$ is a real vector space of dimension $m$ (\cite{2012_lee_SmoothManifolds}).
The Lie algebra $\gothg$ is isomorphic to $\R^m$.
The``wedge" operator $(\cdot)^{\wedge}: \R^m \to \gothg$ is a linear isomorphism from the classical $\R^m$ vector space to the abstract Lie-algebra $\gothg$.
The inverse ``vee" operator $(\cdot)^{\vee}: \gothg \to \R^m$ is defined such that $(v^{\wedge})^{\vee} = v$ for all $v \in \R^m$.

Given $X \in \grpG$ define the left translation $\mathrm L_X : \grpG \to \grpG$ by $\mathrm L_X(Z) = XZ$.
For $U \in \gothg$, the left translation of $U$ by $X$ is defined by $\tD \mathrm L_XU \in \tT_X\grpG$.
Similarly, $\mathrm R_X Z = ZX$ and $\tD \mathrm R_XU \in \tT_X\grpG$ is termed the right translation of $U$ by $X$.
For a matrix Lie group $\tD \mathrm L_XU = XU$ and $\tD \mathrm R_X U = UX$.

For fixed $X \in \grpG$, the adjoint map, $\Ad_X: \gothg \to \gothg$, is defined by
\begin{align*}
\Ad_X(U) &= \tD \mathrm L_X \tD \mathrm R_{X^{-1}} U
\end{align*}
For a matrix Lie group, it is clear that $\Ad_X(U) = X U X^{-1}$.

Given a Lie algebra $\gothg$ and a function $F : \gothg \to \gothg$, $F^{\vee}$ is the corresponding mapping $F^{\vee}: \R^m \to \R^m$ defined by \begin{align*}
F^{\vee}(x) \coloneqq F(x^{\wedge})^{\vee}
\end{align*}
Given a Lie algebra $\gothg$ and a function $H : \R^m \to \gothg$, the notation $H^{\vee} : \R^m \to \R^m$ is defined by
$H^{\vee}(x) \coloneqq H(x)^{\vee}$.
Using analogous notation, log local coordinates around the identity on $\grpG$ are denoted $\log^{\vee} : \grpG \to \R^m$, $\log^\vee (X) := (\log(X))^\vee$.

Let $\calM$ be a smooth manifold and $\grpG$ a Lie group. A \emph{left action} is a function $\phi: \grpG \times \calM \to \calM$ satisfying $\phi(I, \xi) = \xi$ and $\phi(Y, \phi(X, \xi)) = \phi(YX, \xi)$ for all $\xi \in \calM$ and $X, Y \in \grpG$.
For a fixed $X \in \grpG$, the partial map $\phi_X: \calM \to \calM$ maps $\xi \mapsto \phi(X, \xi)$. Similarly, for a fixed $\xi \in \calM$, the partial map $\phi_\xi: \grpG \to \calM$ maps $X \mapsto \phi(X, \xi)$.
A group action is called \emph{free} if for all $\xi \in \calM$, $\phi(X, \xi) = \xi$ only if $X = I$. A group action is called \emph{transitive} if for each pair $\xi, \zeta \in \calM$, there exists a $X \in \grpG$ such that $\phi(X, \xi) = \zeta$.
A \emph{homogeneous space} is a manifold $\calM$ that admits a transitive group action $\phi: \grpG \times \calM \to \calM$. The Lie group $\grpG$ is referred to as a \emph{symmetry} of $\calM$.

If $\grpG$ is a Lie group, the \emph{torsor} $\calG$ is the underlying manifold of $\grpG$ stripped of its group multiplication and group structure.
This gives a natural identification between elements of $\calG$ and $\grpG$.
The notation $\xi \simeq X$ is used to indicate identified elements $\xi \in \calG$ and $X \in \grpG$.
A group torsor naturally inherits a free and transitive left action from the group $\grpG$ induced by applying left translation.
That is, given arbitrary $\xi \in \calG$, $Y \in \grpG$, with $\xi \simeq X \in \grpG$, the group action $\phi : \grpG \times \calG \to \calG$ is defined by $\phi(Y, \xi) \coloneqq \zeta \simeq Y X$.
If multiple Lie groups share the same underlying manifold $\calG$, then this torsor admits multiple symmetries.
We give three examples of this for the state-space of the quadrotor later in this paper.


\section{Problem formulation}
\vspace{-2mm}
\label{sec:Problem formulation}
The following development draws on recent work by the authors (\cite{ARCRAS_Mahony_2022}, \cite{2020_mahony_EquivariantSystems}).
However, in this paper we develop a left-handed symmetry for the control problem, following the lead of \cite{2020_Forbes_Quadrotor} and \cite{2010_Lee_Geometric}, rather than the right-handed symmetry used for the observer problem.
Several results that we use (without proof) in the present development are straightforward analogies of known results for right handed symmetry.

\subsection{Symmetry and Equivariant Systems}
\label{sec:Symmetry}
\vspace{-2mm}

Let $\calM$ be a smooth manifold and the input space $\vecL$ a finite-dimensional vector space.
Consider the dynamical system
\begin{align}
\dot{\xi} = f_u(\xi),  \quad\quad\quad \xi(0) = \xi_0 \label{eq:system_dynamics_manifold}
\end{align}
where $\xi \in \calM$, $u \in \vecL$ and $f: \mathbb{L} \to \gothX(\calM)$ written $u \mapsto f_u$ is affine.
Given a Lie group $\grpG$ acting on the state-space $\calM$, a \emph{lift} (\cite{2020_mahony_EquivariantSystems}) is a smooth function $\Lambda : \calM \times \vecL \to \gothg$ satisfying
\begin{align}
\tD \phi_{\xi} \Lambda(\xi ,u) = f_u(\xi)
\label{eq:lift_def}
\end{align}
where $\phi : \grpG \times \calM \to \calM$ is the group action.
Let $\mr{\xi}$ be a fixed point in $\calM$, termed the origin.
The lifted system is defined to be
\begin{align}
\dot{X}= \tD \mathrm R_{X}\Lambda(\phi_{\mr{\xi}}(X), u) \label{eq:lift}, \quad\quad\quad \phi(X(0), \mr{\xi}) = \xi(0)
\end{align}
for all $X \in \grpG, u \in \vecL$.
A trajectory of the lifted system projects down to a trajectory of the original system via $\phi(X(t), \mr{\xi}) = \xi(t)$ (\cite{2020_mahony_EquivariantSystems}).
Given a suitable group action $\psi: \grpG \times \vecL \to \vecL$, a lift $\Lambda(\xi,u)$ is said to be \emph{equivariant} if it satisfies\footnote{For right-handed symmetry (observer design) the lifted system was $\tD L_{X}\Lambda$ and the equivariant lift condition is for
$\Ad_{X^{-1}} \Lambda(\xi, u) = \Lambda(\phi_X(\xi), \psi(X, u))$ (\cite{ARCRAS_Mahony_2022}).}
\begin{align}
\Ad_{X} \Lambda(\xi, u) &= \Lambda(\phi_X(\xi), \psi(X, u)),
\label{eq:equivariant-lift}
\end{align}
for all $X \in \grpG$, $\xi \in \calM$ and $u \in \vecL$.

In this paper, we are concerned with the particular case where $\calM = \calG$, a Lie-group torsor.
Given that $\calG$ is a group torsor, there is a unique element $I \in \grpG$ that is a natural choice for the reference or origin point, $\mr{\xi} \simeq I$.
Choosing the origin to be the identity element of $\grpG$ allows us to directly identify elements of $\grpG$ and $\calG$ since $\grpG \ni X = XI = \phi_X(I) = X \in \calG$.
The natural left action induced by identification of $\calG$ with $\grpG$ can be used to define the lift $\Lambda(X,u)$ by
\[
\Lambda(X,u) = \tD R_{X^{-1}} f (X,u)
\]
where for $f(X,u)$, $X \in \calG$ is taken in the torsor, while for $\Lambda(X,U)$, $X \in \grpG$ is taken in the group.
Clearly, $f(X,u) =  \tD \mathrm R_{X} \Lambda(X,u)$ and $\Lambda$ is a lift.

In this paper, we consider three different group structures on the same group torsor.
In each case the group multiplication is different and the system lift $\Lambda$ will be different, even though the underlying dynamics $f(\xi,u)$ are the same.
Each lifted system still projects to the same dynamics on the torsor via the different group multiplications.

\subsection{Trajectory tracking on the symmetry group}
\vspace{-2mm}
For the system \eqref{eq:system_dynamics_manifold}, an initial condition $\xi(0) = \xi_0$ along with an input trajectory $u(t)$
for $t \in [0,T]$ determines a \emph{system trajectory}, an integral curve $\xi : [0, T] \to \calM$ satisfying $\dot{\xi} = f_u(\xi)$ for all $t \in [0, T]$.
Given an initial condition $\xi_d(0) = \xi_{d_0}$, and desired input $u_d(t)$ for $t \in [0,T]$, the \emph{desired  trajectory} $\xi_d(t)$ is an integral curve $\xi_d: [0, T] \to \calM$ satisfying  $\dot{\xi}_d = f_{u_d}(\xi_d)$ for all $t \in [0, T]$.
Typically, $\xi_d(t)$ represents a known generated trajectory.

The \emph{tracking} task is to choose a $u(t)$ that steers the system trajectory $\xi(t)$ to $\xi_d(t)$.
In order to formulate the equivariant regulator we identify a lifted trajectory $X_d \in \grpG$ such that $\phi(X_d,\mr{\xi}) = L_{X_d} \mr{\xi} = \xi_d$.
For a general origin $\mr{\xi}$ the different group multiplications lead to different lifted trajectories.
However, by choosing $\mr{\xi} = I$ then $L_{X_d}(\mr{\xi}) = X_d$ for all group structures and we can consider a single lifted system trajectory that is identified with the desired trajectory $X_d \simeq \xi_d$.
This correspondence is only true for free group actions and for the particular choice of origin $\mr{\xi} \simeq I$.

Define the \emph{equivariant error}
\begin{align}
E\coloneqq X_d^{-1}X,
\label{eq:error_E}
\end{align}
noting that $E = I$ if and only if $X$ = $X_d$.
The tracking task is to drive $E \to I$.
The equivariant error will depend on the different group structures chosen and leads to different closed-loop responses.

\section{Equivariant Regulator}
\label{sec:Equivariant Regulator}
\vspace{-2mm}
Let $\calM$ be a manifold and let $\vecL$ be a finite-dimensional vector space. We consider the dynamical system \eqref{eq:system_dynamics_manifold}.
Let $\xi(t)$ be the system trajectory and let $\xi_d(t)$ be a desired trajectory to be tracked.
Let $\grpG$ be a free symmetry on $\calM$ and let $\Lambda$ be a system lift \eqref{eq:lift_def} for $f$.
Then, since the action is free, there exist corresponding trajectories $X(t) =\phi_{\mr{\xi}}^{-1} \xi(t)$, $X_d(t)=\phi_{\mr{\xi}}^{-1} \xi_d(t)$.
Recall the definition \eqref{eq:error_E} of the equivariant error $E$. For simplicity of presentation, we use the group elements $X, X_d \in \grpG$ as the arguments for $\Lambda$ via the identification between $\calM$ and $\grpG$.

\begin{proposition}
	\label{prop:E_dynamics}
	The time derivative of $E$ is given by
	\begin{align}
	\dot{E} = \tD \mathrm R_E \Ad_{X_d^{ -1}}[\Lambda(X_d E, u) - \Lambda(X_d, u_d)].
	\end{align}
\end{proposition}
\vspace{-6mm}
\begin{proof}
	\begin{align*}
	\dot{E}
	&= \tD \mathrm L_{X_d^{ -1}}\dot{X} -\tD \mathrm R_{X} \tD \mathrm L_{X_d^{ -1}}\tD \mathrm R_{X_d^{ -1}}\dot{X_d}\notag\\
	&= \tD \mathrm L_{X_d^{ -1}}\tD \mathrm R_{X}\tD \mathrm R_{X_d^{ -1}} \tD \mathrm R_{X_d}\left[ \Lambda(X, u)\notag -\Lambda(X_d, u_d) \right] \\
	&= \tD \mathrm R_E \Ad_{X_d^{ -1}}[\Lambda(X, u) - \Lambda(X_d, u_d)].\notag
	\end{align*}
	Substituting $X = X_dE$ gives the required result.
\end{proof}
\vspace{-2mm}
If the lift $\Lambda : \calM \times \vecL \to \gothg$ is equivariant, there exists a group action $\psi : \grpG \times \vecL \to \vecL$ such that $\Ad_Y\Lambda(X, u) = \Lambda(YX, \psi(Y, u))$.
In this case, $\dot{E}$ can be written (analogous to \cite{2020_mahony_EquivariantSystems})
\begin{align}
\dot{E} = \tD \mathrm R_E[\Lambda(E, \psi(X_d^{ -1}, u)) - \Lambda(I, \psi(X_d^{ -1}, u_d))]
\end{align}

\begin{definition}\label{def:group_affine}
	If the error dynamics have the form
	\begin{align*}
	\dot{E} = \tD R_E[\Lambda(E, u) - \Lambda(I, u_d)],
	\end{align*}
	then the system is said to be \emph{group affine} (\cite{2017_Bonnabel}).
\end{definition}
Note that $\psi_X \equiv \Id$ implies that the error dynamics are group affine, but that the converse does not hold in general.

\subsection{Error dynamics in local coordinates}
\vspace{-2mm}
The equivariant regulator is based on linearising local coordinates for the error dyanmics.
The local coordinate chart considered is the logarithm $\log^{\vee}: \grpG \to \mathbb{R}^m$ that maps a local neighbourhood of the identity in $\grpG$ to $\gothg$ and then to $\R^m$.
Define notation for the local coordinates $\varepsilon := \log^{\vee}(E)$.

\begin{proposition}
	\label{prop:eps_dynamics}
	The first order dynamics of $\varepsilon$ are given by
	\begin{subequations}\label{eq:linearisation}
	\begin{align}
	\dot{\varepsilon} = A(t)\varepsilon + B(t)\tilde{u} + \mathcal{O}(\| \varepsilon, \tilde{u} \|^2),
	\end{align}
	where
	\begin{align}
	A(t) &= \left(
\Ad_{X_d^{ -1}} \circ \tD_{E | X_d} \Lambda(E, u_d)
	\circ \tD \mathrm L_{X_d}
	\right)^{\vee} \label{eq:lin_A},
	\\
	B(t) &= \left(
	\Ad_{X_d^{ -1}}
	\circ \tD_{u | u_d} \Lambda(X_d, u)
	\right)^{\vee} \label{eq:lin_B}.
	\end{align}
If $f$ is equivariant, then the system matrices can be simplified to
	\begin{align}
	A(t) &= \left(\tD_{E |  I}\Lambda(E, \psi(X_d^{ -1}, u_d))\right)^{\vee} \label{eq:lin_A_eq},  \\
	B(t) &= \left( \tD_{u | \psi(X_d^{ -1}, u_d)}\Lambda(\Id, u)\tD_{u | u_d}\psi(X_d^{ -1}, u) \right)^{\vee} \label{eq:lin_B_eq}.
	\end{align}

If $f$ is group affine, then
\begin{align}
A(t) &= \left(\tD_{E |  I}\Lambda(E, u_d)\right)^{\vee} \label{eq:lin_A_GA}, \\
B(t) &= \left( \tD_{u | u_d}\Lambda(\Id, u) \right)^{\vee} \label{eq:lin_B_GA}.
\end{align}
\end{subequations}
\end{proposition}
The proof can be found in Appendix \ref{sec:Proof4.4}.

\section{Example}
\label{sec:quadrotor_model}
\vspace{-2mm}
In this section we consider trajectory tracking control for a quadrotor vehicle.
The quadrotor is modelled as a rigid body, with state $\xi = (R, x, v) \in \mathcal{SO}(3) \times \mathbb{R}^3 \times \mathbb{R}^3$.
The matrix $R$ denotes the orientation of the body frame with respect to the inertial frame, and $x$ and $v$ respectively denote the position and translational velocity of the body frame with respect to the inertial frame, expressed in the inertial frame.

The input $u = \left( \begin{matrix} \Omega, T \end{matrix} \right) \in \vecL = \R^4$ consists of the body-fixed angular velocity $\Omega \in \mathbb{R}^3$ and the combined rotor thrust $T \in \mathbb{R}$.
There are low-level angular rate and rotor speed control loops that we do not model in the present paper.
The skew symmetric operator $(\cdot)^{\times}: \mathbb{R}^3 \to \so(3)$ (skew-symmetric matrices) is defined such that $a^{\times}b = a \times b$ for all $a, b \in \mathbb{R}^3$.
The quadrotor dynamics,
$
\dot{\xi} = f_{u}(\xi) = (\dot{R}, \dot{x}, \dot{v})
$
for $f : \vecL \to \gothX(\mathcal{SO}(3) \times \R^3 \times \R^3)$ are given by (\cite{2020_Forbes_Quadrotor})
\begin{subequations}\label{eq:quadrotor_dynamics_orig}
\begin{align}
\dot{R} &= R\Omega^{\times}, \\
\dot{x} &= v,   \\
\dot{v} &= -\frac{T}{m}Re_3 + g e_3 - RD R^\top v,
\end{align}
\end{subequations}
where the matrix $D$ denotes rotor drag, $m$ denotes the mass of the quadrotor and $g$ denotes the acceleration due to gravity.

For this paper, we shall assume that wind speed is negligible. We follow the common approach taken in the literature (\cite{Mahony_quadrotor,kai_2017}) and model aerodynamic drag as a linear drag in the rotor plane \cite[Eq.~10]{Mahony_quadrotor}.
That is,
\begin{align}
D & = c_1 (I - e_3e_3^\top).
\end{align}
Define a new input, termed the compensated thrust (\cite{kai_2017}) by
\begin{align}
\bar{T} = \frac{1}{m} T - c_1 e_3^\top  R^\top v \label{eq:thrust_modification}.
\end{align}
Setting the input $u = \left( \begin{matrix} \Omega, \bar{T} \end{matrix} \right) \in \vecL = \R^4$, the resulting dynamics are
\begin{subequations} \label{eq:quadrotor_dynamics}
\begin{align}
\dot{R} &= R\Omega^{\times}, \\
\dot{x} &= v,   \\
\dot{v} &= -\bar{T}Re_3 + g e_3 - c_1 v. \label{eq:quad_velocity}
\end{align}
\end{subequations}
This input transformation significantly simplfies the trajectory generation (\S\ref{sec:trajectory_generation}) and the symmetry analysis that follows.

The manifold $\mathcal{SO}(3) \times \mathbb{R}^3 \times \mathbb{R}^3$ serves as the torsor for the following three distinct Lie group structures.

\subsubsection{Direct product symmetry:}
\label{sub:direct_product}
\vspace{-2mm}
The first realisation considered uses the direct product group structure
$\SO(3) \times \R^3 \times \R^3$ where $\SO(3)$ is the special orthogonal group and $\R^3$ is a group under addition.
The group product is given by
\begin{align*}
(Q_1, p_1, \gamma_1)(Q_2, p_2, \gamma_2) = (Q_1Q_2, p_1 + p_2, \gamma_1 + \gamma_2).
\end{align*}
The Lie algebra associated with $\SO(3) \times \R^3 \times \R^3$ is $\so(3) \times \R^3 \times \R^3$.
The group action on $\cal{SO}(3) \times \R^3 \times \R^3$ is derived from identifying the system coordinates with group elements:
$\phi((Q,p,\gamma), (R,x,v)) = (QR, p + x, \gamma + v)$.
This action is free and transitive.

Comparing \eqref{eq:quadrotor_dynamics} to the definition \eqref{eq:lift}, the system lift $\Lambda : (\SO(3) \times \R^3 \times \R^3) \times \vecL \to (\gothso(3) \times \R^3 \times \R^3)$ is given by
\begin{subequations} \label{eq:so3_lin}
\begin{align}
&\Lambda_R = R\Omega^\times R^\top,\\
&\Lambda_x = v,\\
&\Lambda_v = -\bar{T}Re_3 + ge_3 - c_1 v, \label{so3 lift}
\end{align}
\end{subequations}
where $\Lambda((R, x, v), u) = (\Lambda_R, \Lambda_x, \Lambda_v)$.
The induced equivariant error is
\begin{align*}
E &= (R_d^\top R, x - x_d, v - v_d ).
\end{align*}

\subsubsection{Extended Pose symmetry:}
\label{sec:se23_equivariant}
\vspace{-2mm}
The extended pose symmetry $\SE_2(3)$ integrates a special Euclidean symmetry on position and velocity (\cite{2014_Bonnabel}).
Identifying the system coordinates with group elements, the group action on $\cal{SO}(3) \times \R^3 \times \R^3$ is given by
\begin{align*}
\phi((Q,p,\gamma), (R,x,v)) = (QR, Qx + p, Qv + \gamma).
\end{align*}
The corresponding system lift $\Lambda : \SE_2(3) \times \vecL \to \se_2(3)$ is given by
\begin{subequations} \label{eq:se23 lift}
\begin{align}
&\Lambda_R = R\Omega^\times R^\top, \\
&\Lambda_x = -R\Omega^{\times}R^\top x + v,\\
&\Lambda_v = -R\Omega^{\times}R^\top v - \bar{T}Re_3 + ge_3 -c_1v,
\end{align}
\end{subequations}
where $\Lambda((R, x, v), u) = (\Lambda_R, \Lambda_x, \Lambda_v) \in \se_2(3)$.
The equivariant error is defined by
\begin{align*}
E &= (R_d^\top R, R_d^\top (x - x_d), R_d^\top (v - v_d)).
\end{align*}
By computing $(\Lambda(E, u) - \Lambda(I, u_d))E$, the lifted quadrotor system on extended pose $\SE_2(3)$ can be shown to be group affine (Definition \ref{def:group_affine}).
This also marks an important distinction between the development here and the development in \cite{2020_Forbes_Quadrotor}, in which the quadrotor system model on $\SE_2(3)$ is not group affine.
It is interesting to note that the direct product lifted system (\ref{sub:direct_product}) is not group affine: in this sense the extended pose lifted system is more structured.

\subsubsection{Pose and velocity symmetry:}
\vspace{-2mm}
The pose and velocity symmetry treats the attitude and position as a pose with an $\SE(3)$ (special Euclidean group) symmetry and uses a direct product $\R^3$ symmetry for velocity.
The group product is given by
\begin{align*}
(Q_1,  p_1, \gamma_1 ) (Q_2, p_2, \gamma_2)
=
(Q_1 Q_2 , Q_1p_2 + p_1, \gamma_1 + \gamma_2).
\end{align*}
The group action on $\cal{SO}(3) \times \R^3 \times \R^3$ is given by
\begin{align*}
\phi((Q,p,\gamma), (R,x,v)) = (QR, Qx + p, v + \gamma).
\end{align*}

Since the velocity is treated separately, it is possible to model either the inertial velocity or the body-frame velocity.
The system model \eqref{eq:quadrotor_dynamics} used an inertial velocity.
However, setting $v_b = R^\top v$ and replacing \eqref{eq:quad_velocity} by $\dot{v}_b =-\Omega^\times v_b -\frac{1}{m}\bar{T}e_3 + gR^\top e_3 - c_1 v_b$ the resulting system is written in terms of a body velocity. That is, the new system state is written $\xi = (R, x, v_b)$ where $v = R v_b$.
This change allows us to find a finite-dimensional equivariant extension for the system model that was not possible for the standard model with the pose and velocity symmetry.

The corresponding lift $\Lambda : (\SE(3) \times \R^3) \times \vecL \to \se(3) \times \R^3$ is given by
\begin{subequations} \label{eq:se3_lift}
\begin{align}
&\Lambda_R = R\Omega^\times R^\top, \\
&\Lambda_x = -R\Omega^\times R^\top x + Rv_b, \\
&\Lambda_{v_b} = -\Omega^\times v_b -\bar{T}e_3 + R^\top ge_3 - c_1 v_b,
\end{align}
\end{subequations}
where $\Lambda((R, x, v), u) = (\Lambda_R, \Lambda_x, \Lambda_{v_b}) \in \se(3) \times \R^3$.
The resulting system can be shown to be equivariant with an appropriate velocity extension (see Appendix \ref{sec:Equivariance SE(3)}).
The system, however, is not group affine.

The equivariant error is given by
$E = (R_d^\top R, R_d^\top (x - x_d), v - v_d)$.

\section{Simulation}
\label{sec:simulation}
\vspace{-2mm}
\subsection{Trajectory Generation}
\label{sec:trajectory_generation}
\vspace{-2mm}
The quadrotor system \eqref{eq:quadrotor_dynamics} is differentially flat (\cite{2011_min_snap_kumar}). We choose the flat outputs to be the position $x_d \in \R^3$ and the heading $\alpha_d \in [0, 2\pi)$.

Given $(x_d(t), \alpha_d(t))$, $v_d$ is given directly by $\dot{x}_d$.
Then, from the quadrotor dynamics \eqref{eq:quadrotor_dynamics}, $R_de_3$ and $\bar{T}_d$ can be computed via
\begin{equation*}
R_de_3 = \frac{-\ddot{x}_d  + ge_3 - c_1v_d}{\lVert-\ddot{x}_d  + ge_3 - c_1v_d \rVert} \mbox{ and } \bar{T}_d = \lVert-\ddot{x}_d  + ge_3 - c_1v_d \rVert.
\end{equation*}
Defining $\theta = \cos^{-1}\left( e_3^\top R_de_3 \right)$, $\mu = e_3 \times (R_de_3)$, the full attitude $R_d$ can be computed by $R_d = \exp(\alpha_d e_3)\exp(\theta \mu)$.
The angular velocity $\Omega_d$ can be computed via the time derivative of the attitude, $\Omega_d^{\times} = R_d^{-1}\dot{R}_d$.
Note that in contrast to \cite{2019_Farrell_Error} and \cite{2020_Forbes_Quadrotor}, there is no requirement to neglect drag forces or compute an iterative solution to compute $R_d$ due to the thrust modification \eqref{eq:thrust_modification}.
\vspace{-2mm}
\subsection{Symmetry group linearisations of the quadrotor system}
\label{sec:quadrotor_linearisations}
\vspace{-2mm}
Applying the methodology outlined in Section~\ref{sec:Equivariant Regulator} to the quadrotor systems discussed in Section~\ref{sec:quadrotor_model}, we obtain the following linearisations (in local logarithmic coordinates). In all cases, $\varepsilon = \begin{pmatrix} r_{\varepsilon}, x_{\varepsilon}, v_{\varepsilon} \end{pmatrix}$ and $\tilde{u} = \left( \begin{matrix} \Omega - \Omega_d, \bar{T} - \bar{T}_d \end{matrix}\right)$.

\textbf{Direct product symmetry} ($\SO(3) \times \R^3 \times \R^3$):
\begin{align}
\dot{\varepsilon} &= \left(\begin{matrix}
-\Omega_d^{\times} & 0 & 0 \\
0 & 0 & I\\
\frac{\bar{T}_d}{m}R_de_3^{\times} & 0 & -c_1 I
\end{matrix} \right) \varepsilon + \left(\begin{matrix}
I & 0 \\
0 & 0\\
0 & -R_de_3
\end{matrix} \right)\tilde{u}.
\label{eq:AB_direct_product}
\end{align}

\textbf{Extended pose symmetry} ($\SE_2(3)$):
\begin{align}
\dot{\varepsilon}= \left(\begin{matrix}
-\Omega_d^{\times} & 0 & 0\\
0 & -\Omega_d^{\times} & I \\
\frac{\bar{T}_d}{m}e_3^{\times} & 0 & -\Omega_d^{\times} - c_1 I\\ \end{matrix}\right)\varepsilon +
\left(\begin{matrix}
I & 0\\
0 & 0 \\
0 & -e_3\\ \end{matrix}\right)
\tilde{u}. \label{eq:se23_linearised}
\end{align}

\textbf{Pose and velocity symmetry} ($\SE(3) \times \R^3$):
\begin{align}
\dot{\varepsilon} = \left(\begin{matrix}
-\Omega_d^\times & 0 & 0\\
-v_{b_d}^\times & -\Omega_d^\times & I \\
g(R_d^\top e_3)^\top & 0 & -\Omega_d^\times -c_1I\\ \end{matrix}\right)\varepsilon +
\left(\begin{matrix}
I & 0\\
0 & 0 \\
v_{b_d}^\times & -e_3\\ \end{matrix}\right)
\tilde{u}.
\label{eq:AB_pose_velocity}
\end{align}

\subsection{Simulation Results}
\vspace{-2mm}
A simple case study is presented to compare the relative performance of the EqR design using different symmetry groups.
The trajectories are designed and tracked by simulating the quadcopter and controller dynamics in Python.
The quadrotor states are assumed to be known exactly at all times.
The drag constant $c_1 = 0.25$ is used and the discrete-time LQR parameters are as follows:
$r_\varepsilon$ state cost matrix $1.0 I_3$;
$x_\varepsilon$ state cost matrix $2.0 I_3$;
$v_\epsilon$ state cost matrix $0.1 I_3$;
$\tilde{\Omega}$ input cost matrix $0.5 I_3$;
and $\tilde{\bar{T}}$ scalar input cost $0.5$.

\subsection{Point regulation and transient response}
\label{sec:MonteCarlo}
\vspace{-2mm}
The goal of this simulation is to investigate the transient responses of the different controllers when tracking a hover point trajectory, defined by
\begin{equation*}
x_d(t)= (0, 0, 0)  \mbox { and } \alpha_d(t) = 0, \mbox{ for } t \in [0, \pi].
\end{equation*}
This trajectory has the property that the linearisations (\ref{eq:AB_direct_product}-\ref{eq:AB_pose_velocity}) at the goal trajectory are identical.
Thus, the only differences in response will be due to transient effects.

\begin{figure}	\includegraphics[width=0.6\linewidth]{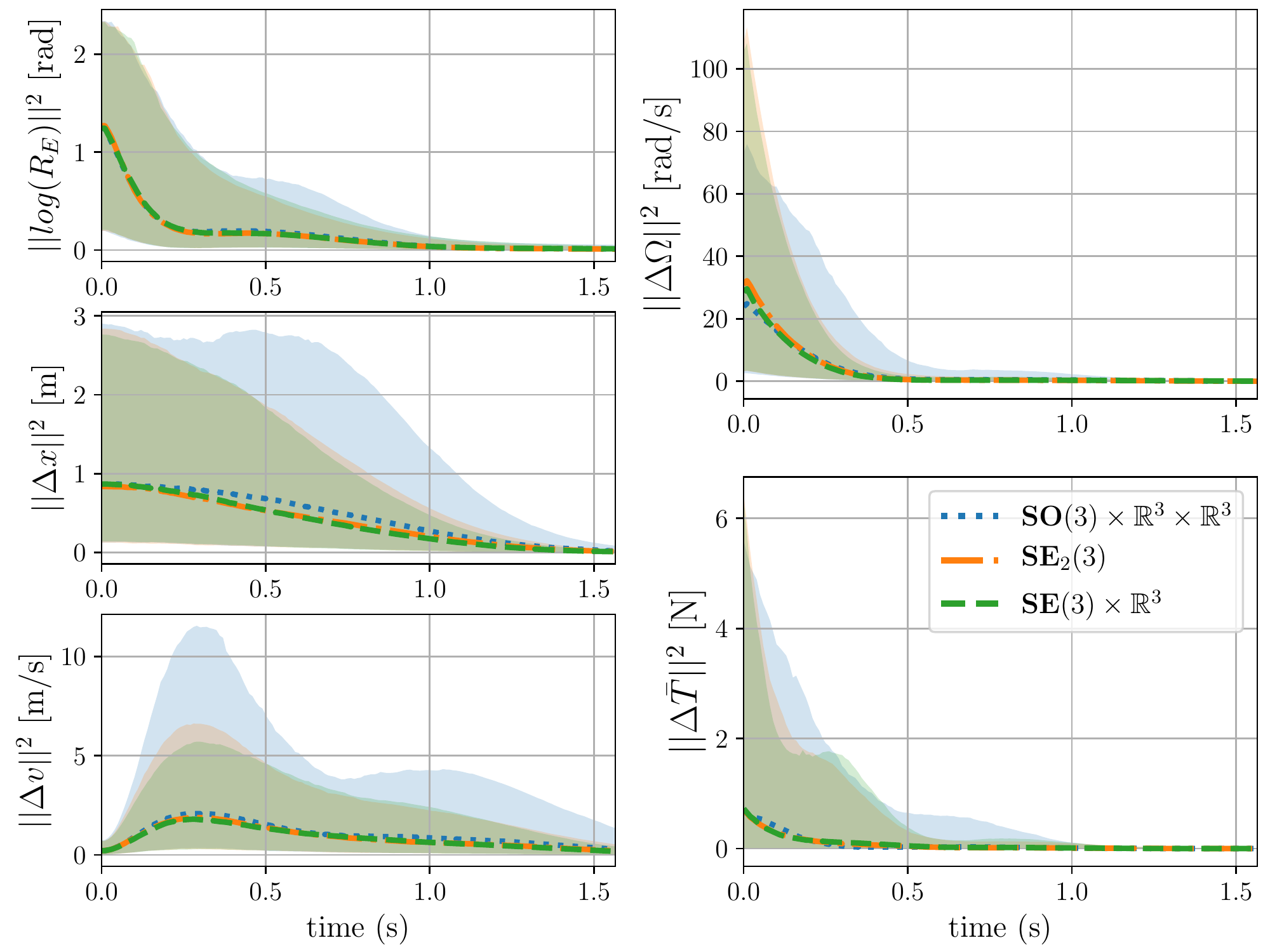}
	\centering
	\caption{ Median squared norm of state error regulating hover over time for 2000 simulations. Convergence is obvious after $\frac{\pi}{2}$ seconds, so only results for $t \in [0, \frac{\pi}{2}]$ are shown. Shaded area shows 5th to 95th percentile of mean squared error.}
	\label{fig:hover_IC}
\end{figure}

The initial orientation, position and velocity are randomly perturbed in each component by sampling from Gaussian distributions of means $0$ and standard deviations 0.8, 0.6 and 0.3, respectively.
The results of 2000 Monte Carlo trials are shown in Figure~\ref{fig:hover_IC} and discussed in \S~\ref{sec:Dicussion}.

\subsection{Lissajous trajectory and transient response}
\begin{figure}
\includegraphics[width=0.6\linewidth]{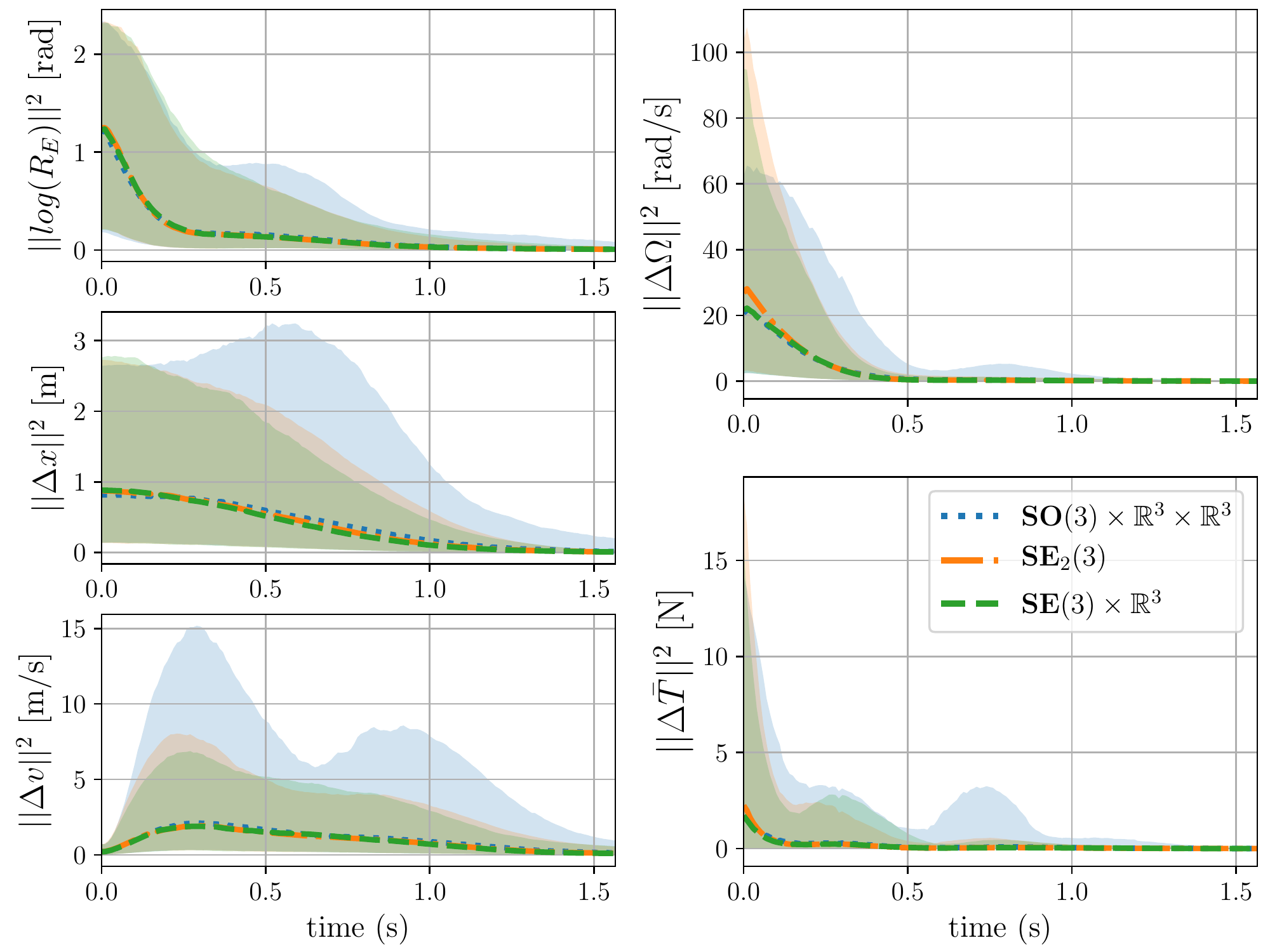}
	\centering
	\caption{ Median squared norm of state error tracking Lissajous curve over time for 2000 simulations. Convergence is obvious after $\frac{\pi}{2}$ seconds, so only results for $t \in [0, \frac{\pi}{2}]$ are shown. Shaded area shows 5th to 95th percentile of mean squared error.
	}
	\label{fig:lissa_IC}
\end{figure}

\vspace{-2mm}
The goal of this simulation is to investigate the transient responses of the different controllers when tracking a Lissajous trajectory, defined by
\begin{equation}
x_d(t)=  \frac{1}{2}(\cos(2t),\sin(4t), 2)  \mbox {, } \alpha_d(t) = 0 \mbox{, } t \in [0, \pi]. \label{eq:lissajous}
\end{equation}
This trajectory has the property that is non-trim, and so will result in time-varying $A$ and $B$ matrices for all symmetries.
The initial orientation, position and velocity are randomly perturbed in each component by sampling from Gaussian distributions of means $0$ and standard deviations 0.8, 0.6 and 0.3, respectively.
The results of 2000 Monte Carlo trials are shown in Figure~\ref{fig:lissa_IC} and discussed in \S~\ref{sec:Dicussion}.

\subsection{Asymptotic tracking performance}
\begin{figure}
	\includegraphics[width=0.6\linewidth]{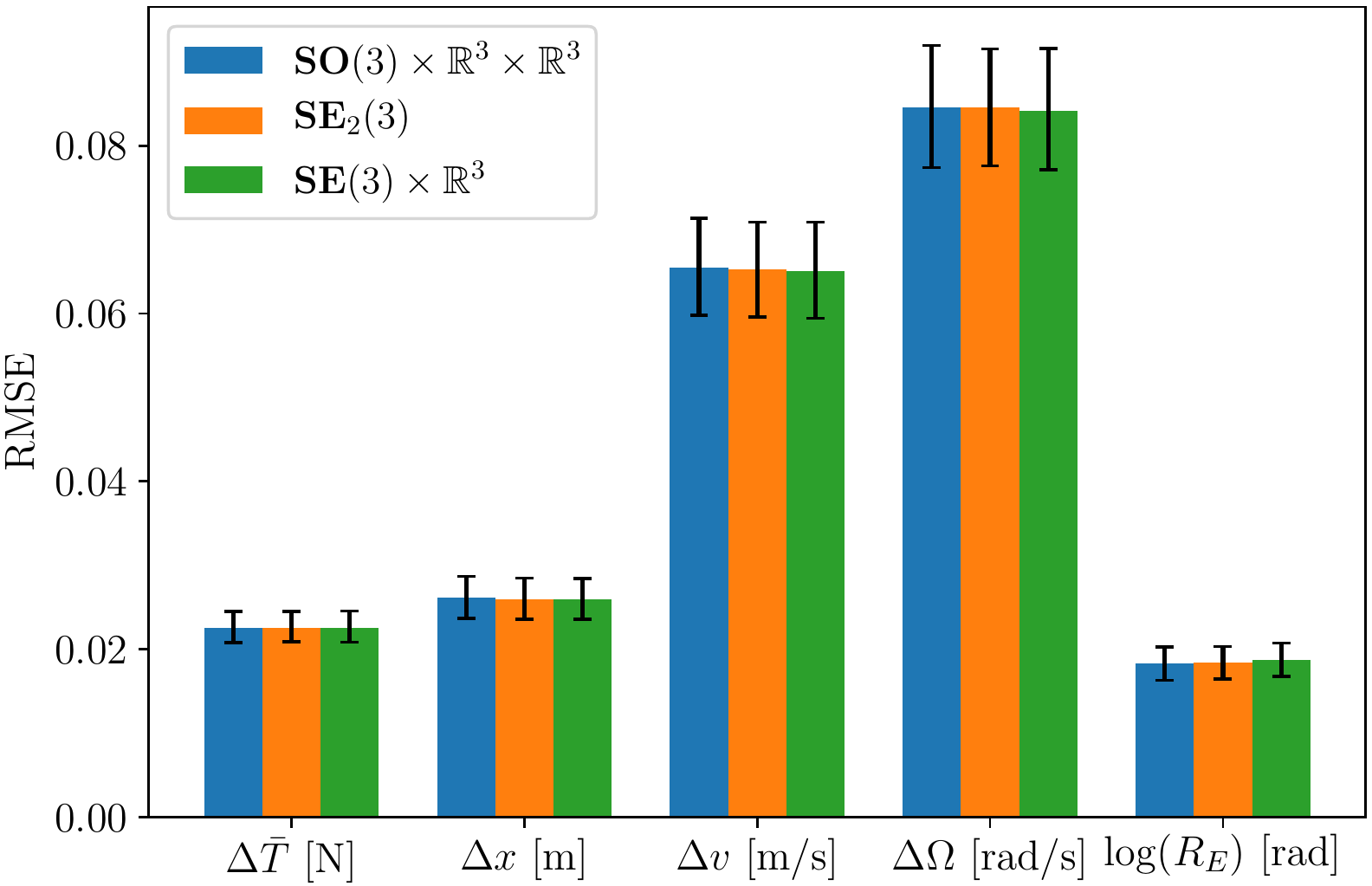}
	\centering
	\caption{ RMSE tracking Lissajous curve over time for 5000 simulations in the presence of process noise. Bar shows median and error bars represent 20th and 80th percentile.}
	\label{fig:lissa_rmse}
\end{figure}

\vspace{-2mm}
The goal of this simulation is to investigate the asymptotic responses of the different controllers.
The trajectory to be tracked is the Lissajous curve \eqref{eq:lissajous}.
The initial states are exactly the initial desired states, but the system derivatives are perturbed in each component at every time step by sampling from Gaussian distributions of mean $0$ and standard deviation $0.1$.
The distribution of RMSE over 5000 Monte Carlo trials are shown in Figure~\ref{fig:lissa_rmse} and discussed in \S~\ref{sec:Dicussion}.

\subsection{Discussion}
\label{sec:Dicussion}
\vspace{-2mm}
There are two important factors to analyse in Figure~\ref{fig:hover_IC}: the median response and the outlier performance.
The median response of each controller is very similar, although there is a higher error seen in the transient response of the direct product controller in the position component.
There is a significantly wider range of performance of the direct product controller compared to the extended pose and pose and velocity controllers across all state and input components; that is, the direct product controller performs poorly more frequently.
Importantly, the performance discrepancy is not simply due to the controllers expending more energy; indeed, the direct product symmetry appears to consistently expend more energy for worse performance.
The results for tracking the Lissajous curve with different initial conditions in Figure~\ref{fig:lissa_IC} are almost identical to the hover regulation results in Figure \ref{fig:hover_IC}.
This indicates that the observed performance of the controllers is not trajectory-specific.
Figure~\ref{fig:lissa_rmse} reveals no discernible difference in asymptotic tracking performance between symmetries.

These simulations, along with the fact that the hover trajectory results in identical linearised dynamics and error definitions, indicate that the salient differences must be due to the particular symmetry's Lie algebraic representation of the system state, and the associated higher-order system dynamics. 
Previous research has shown that choices of equivariant or group affine symmetries lead to lower linearisation error (\cite{Bonnabel_2019}, \cite{Mahony_2022}).
A possibility for the improved performance seen here is that the group affine structure of the extended pose and the equivariant structure of the pose and velocity symmetry have reduced linearisation error.
In this way, identifying specific properties of a given symmetry reveals insight into the underlying geometry of the system and provides motivation for choice of symmetry prior to implementation.

The improved performance of the extended pose symmetry compared to the direct product symmetry is consistent with the results of \cite{2020_Forbes_Quadrotor}. This, and the improved performance of the pose and velocity symmetry compared to the direct product symmetry suggests that the most important component of rigid body symmetries is the coupling of the rotational and positional components of the state.
The quadrotor dynamics are otherwise invariant with respect to position, so this coupling may positively impact the LQR Riccati equation solutions (for these trajectories and sets of LQR gains).

\section{Conclusion} \label{sec:conclusion}
\vspace{-2mm}
This paper presented the Equivariant Regulator: a general methodology for trajectory tracking on manifolds with a free symmetry.
We discussed several special cases of symmetry, and analysed how these special cases simplify error trajectory linearisation.
We presented the sample case of the quadrotor, generated sample trajectories and used the methodology to derive error trajectory linearisations and LQR controllers to track these trajectories in the presence of perturbations. We showed that equivariant and group affine symmetries result in improved performance in the presence of initial perturbations.

\bibliographystyle{plainnat}
\bibliography{references}             
\vspace{-2mm}                                                     







\newpage

\appendix
\section{Proof of Proposition \ref{prop:eps_dynamics}}
\label{sec:Proof4.4}

\begin{proof}[Proof of \ref{eq:lin_A} and \ref{eq:lin_B}]
	The dynamics of $\varepsilon$ are given by
	\begin{align}
	\dot{\varepsilon} &= D\log^{\vee}(E)\dot{E} \notag\\
	&= D\log^{\vee}(E)\tD R_E\Ad_{X_d^{-1}}[\Lambda(X_dE, u) - \Lambda(X_d, u_d)] \notag\\
	&= D\log^{\vee}(\exp(\varepsilon^{\wedge}))\tD R_{\exp(\varepsilon^{\wedge})}\Ad_{X_d^{-1}}[\Lambda(X_d\exp(\varepsilon^{\wedge}), u) \notag\\
	&\quad \quad \quad \quad \quad \quad \quad \quad \quad \quad \quad \quad \quad -\Lambda(X_d, u_d)] \label{eq:epsilon_def}
	\end{align}
	
	Substituting $u = \tilde{u} + u_d$, and defining
	\begin{align*}
	F(\varepsilon) = D\log^{\vee}(\exp(\varepsilon^{\wedge}))\tD R_{\exp(\varepsilon^{\wedge})}
	\end{align*}
	$\dot{\varepsilon}$ becomes
	\begin{align}
	\dot{\varepsilon}=F(\varepsilon)\Ad_{X_d^{-1}}[\Lambda(X_d\exp(\varepsilon^{\wedge}), \tilde{u} + u_d)-\Lambda(X_d, u_d)] \label{eq:eps_dot}
	\end{align}
	
	Note that $\dot{\varepsilon}$ is a function of $\varepsilon$ and $\tilde{u}$ and so the notation of $\dot{\varepsilon}$ will briefly be overloaded to make this explicit: $\dot{\varepsilon} = \dot{\varepsilon}(\varepsilon, \tilde{u})$.
	
	By definition, the Taylor expansion of $\dot{\varepsilon}(\varepsilon, \tilde{u})$ around $\varepsilon = 0, \tilde{u} = 0$ is given by
	\begin{align*}
	\dot{\varepsilon}(\varepsilon, \tilde{u}) = \dot{\varepsilon}(0, 0)
	+ \tD_{\varepsilon | 0 }\dot{\varepsilon}(\varepsilon, 0)[\varepsilon]
	+ \tD_{\tilde{u} | 0}\dot{\varepsilon}(0, \tilde{u})[\tilde{u}]
	\\+ \mathcal{O}(\| \varepsilon, \tilde{u} \|^2 )
	\end{align*}
	This expression will be approached term-by-term. Firstly,
	\begin{align*}
	\dot{\varepsilon}(0, 0) =F(0)\Ad_{X_d^{-1}}[\Lambda(X_d,u_d)-\Lambda(X_d, u_d)] = 0
	\end{align*}
	
	Next, the first-order expansion in $\varepsilon$.
	\begin{align}
	\tD_{\varepsilon |  0 }&\dot{\varepsilon}(\varepsilon, 0)[\varepsilon] =\notag\\ &\tD_{\varepsilon | 0 } F(\varepsilon)\Ad_{X_d^{-1}}[\Lambda(X_d\exp(\varepsilon^{\wedge}), u_d)\notag \\
	&\quad \quad\quad\quad\quad\quad\quad\quad\quad\quad-\Lambda(X_d, u_d)] [\varepsilon]\notag \\
	=&\tD_{\varepsilon | 0}(F(\varepsilon))[\varepsilon]\cdot\Ad_{X_d^{-1}}[\Lambda(X_d, u_d)-\Lambda(X_d, u_d)]\notag\\
	&\phantom{=}+F(0)\Ad_{X_d^{-1}}\tD_{\varepsilon|0}[\Lambda(X_d\exp(\varepsilon^{\wedge}), u_d)\notag \\
	&\quad \quad\quad\quad\quad\quad\quad\quad\quad\quad-\Lambda(X_d, u_d)][\varepsilon] \label{eq:product_rule}\\
	=& \left( \Ad_{X_d^{-1}}\tD_{\varepsilon | 0}\Lambda(X_d\exp(\varepsilon^{\wedge}),  u_d)\right)^{\vee} [\varepsilon]\notag\\
	=& \left(\Ad_{X_d^{-1}}\tD_{E | X_d}\left(\Lambda(E, u_d)\right)\tD L_{X_d} \right)^{\vee} [\varepsilon]
	\end{align}
	
	where the product rule has been used in \eqref{eq:product_rule}.
	
	Finally, the first order expansion in $\tilde{u}$.
	
	\begin{align}
	\tD_{\tilde{u} | 0}&\dot{\varepsilon}(\varepsilon, \tilde{u})[\tilde{u}] =\notag\\ & F(0)\Ad_{X_d^{-1}}\tD_{\tilde{u} |  0}\Lambda(X_d\exp(0^{\wedge}), \tilde{u} + u_d) [\tilde{u}]\notag\\
	=& \left( \Ad_{X_d^{-1}}\tD_{u | u_d}\Lambda(X_d, u) \right)^{\vee} [\tilde{u}]\notag
	\end{align}
\end{proof}

\begin{proof}[Proof of \ref{eq:lin_A_eq} and \ref{eq:lin_B_eq}]
	$A(t)$ and $B(t)$ can first be derived as above. Then, due to equivariance of $\Lambda$, we have
	\begin{align*}
	A(t) = \left( \Ad_{X_d^{-1}}\tD_{E | X_d}\left(\Lambda(e, u_d)\right)\tD L_{X_d} \right)^{\vee}\\
	= \left( \tD_{E |  X_d }\left(\Lambda(X_d^{-1}E, \psi(X_d^{-1}, u_d))\right)\tD L_{X_d} \right)^{\vee}\\
	= \left( \tD_{E | I }\left(\Lambda(E, \psi(X_d^{-1}, u_d))\right)\tD L_{X_d^{-1}}DL_{X_d} \right)^{\vee}\\
	= \left(\tD_{E | I}\Lambda(E, \psi(X_d^{-1}, u_d))\right)^{\vee}
	\end{align*}
	and
	\begin{align*}
	B(t) = \left(\Ad_{X_d^{-1}} \tD_{u | u_d}\Lambda(X_d, u) \right)^{\vee}\\
	= \left( \tD_{u | u_d}\Lambda(X_d^{-1}X_d, \psi(X_d^{-1}, u)) \right)^{\vee}\\
	= \left( \tD_{u | \psi(X_d^{ -1}, u_d)}\Lambda(I,u)\tD_{u | u_d}\psi(X_d^{-1}, u) \right)^{\vee}
	\end{align*}
\end{proof}

\begin{proof}[Proof of \ref{eq:lin_A_GA} and \ref{eq:lin_B_GA}]
	From definition \ref{def:group_affine} and the definition of $\varepsilon$, the dynamics of $\varepsilon$ are given by
	\begin{align*}
	\dot{\varepsilon} &= D\log^{\vee}(E)\dot{E} \notag\\
	&= D\log^{\vee}(E)\tD R_E[\Lambda(E, u) - \Lambda(I, u_d)] \notag\\
	&= D\log^{\vee}(\exp(\varepsilon^{\wedge}))\tD R_{\exp(\varepsilon^{\wedge})}[\Lambda(\exp(\varepsilon^{\wedge}), u) \notag\\
	&\quad \quad \quad \quad \quad \quad \quad \quad \quad \quad \quad \quad \quad -\Lambda(I, u_d)]
	\end{align*}
	Substituting $u = \tilde{u} + u_d$, and defining
	\begin{align*}
	F(\varepsilon) = D\log^{\vee}(\exp(\varepsilon^{\wedge}))\tD R_{\exp(\varepsilon^{\wedge})}
	\end{align*}
	$\dot{\varepsilon}$ becomes
	\begin{align}
	\dot{\varepsilon}=F(\varepsilon)[\Lambda(\exp(\varepsilon^{\wedge}), \tilde{u} + u_d)-\Lambda(I, u_d)]
	\end{align}
	By definition, the Taylor expansion of $\dot{\varepsilon}(\varepsilon, \tilde{u})$ around $\varepsilon = 0, \tilde{u} = 0$ is given by
	\begin{align*}
	\dot{\varepsilon}(\varepsilon, \tilde{u}) = \dot{\varepsilon}(0, 0)
	+ \tD_{\varepsilon | 0 }\dot{\varepsilon}(\varepsilon, 0)[\varepsilon]
	+ \tD_{\tilde{u} | 0}\dot{\varepsilon}(0, \tilde{u})[\tilde{u}]
	\\+ \mathcal{O}(\| \varepsilon, \tilde{u} \|^2 )
	\end{align*}
	Again approaching this sum term-by-term,
	\begin{align*}
	\dot{\varepsilon}(0, 0) =F(0)[\Lambda(I,u_d)-\Lambda(I, u_d)] = 0
	\end{align*}
	Next, the first-order expansion in $\varepsilon$.
	\begin{align}
	\tD_{\varepsilon |  0 }&\dot{\varepsilon}(\varepsilon, 0)[\varepsilon] =\notag\\ &\tD_{\varepsilon | 0 } F(\varepsilon)[\Lambda(\exp(\varepsilon^{\wedge}), u_d)\notag \\
	&\quad \quad\quad\quad\quad\quad\quad\quad\quad\quad-\Lambda(X_d, u_d)] [\varepsilon]\notag \\
	=&\tD_{\varepsilon | 0}(F(\varepsilon))[\varepsilon]\cdot[\Lambda(I, u_d)-\Lambda(I, u_d)]\notag\\
	&\phantom{=}+F(0)\tD_{\varepsilon|0}[\Lambda(\exp(\varepsilon^{\wedge}), u_d)\notag \\
	&\quad \quad\quad\quad\quad\quad\quad\quad\quad\quad-\Lambda(I, u_d)][\varepsilon] \label{eq:product_rule2}\\
	=& \left( \tD_{\varepsilon | 0}\Lambda(\exp(\varepsilon^{\wedge}),  u_d)\right)^{\vee} [\varepsilon]\notag\\
	=& \left(\tD_{E | I}\left(\Lambda(E, u_d)\right) \right)^{\vee} [\varepsilon]
	\end{align}
	Finally, the first order expansion in $\tilde{u}$.
	
	\begin{align}
	\tD_{\tilde{u} | 0}&\dot{\varepsilon}(\varepsilon, \tilde{u})[\tilde{u}] =\notag\\ & F(0)\tD_{\tilde{u} |  0}\Lambda(\exp(0^{\wedge}), \tilde{u} + u_d) [\tilde{u}]\notag\\
	=& \left( \tD_{u | u_d}\Lambda(I, u) \right)^{\vee} [\tilde{u}]\notag
	\end{align}
\end{proof}
\section{Equivariance of $\SE(3) \times \R^3$ system}
\label{sec:Equivariance SE(3)}
To see the equivariance of the quadrotor system on $\SE(3) \times \R^3$, define the following \emph{extended input}
\begin{align*}
u^* = \left( \begin{matrix} \Omega, \bar{T}^{*}, g^{*}, w \end{matrix} \right) \in \R^{12} = \vecL^*
\end{align*}
and consider the \emph{extended system lift} $\Lambda : \SE_2(3) \times \vecL^* \to \se_2(3)$ defined by

\begin{align*}
&\Lambda_R = R\Omega^\times R^\top \notag\\
&\Lambda_x = -R\Omega^\times R^\top x + R(v_b - w) \notag\\
&\Lambda_{v_b} = -\Omega^\times (v_b - w) -\frac{1}{m}\bar{T}^* + R^\top g^* - c_1 (v_b - w)
\end{align*}

Noting that the lift \eqref{eq:se3_lift} is captured by setting $w = 0, \bar{T}^* = \bar{T}e_3$ and $g^* = ge_3$.

Define the left input action $\psi : \SE_2(3) \times \vecL^* \to \vecL^*$ by
\begin{align*}
\psi_Y:
\left(\begin{matrix}
\Omega \\
\bar{T}^*\\
g^*\\
w  \end{matrix}\right) \mapsto \left(\begin{matrix}
\Omega \\
\bar{T}^*\\
R_Y g^*\\
w + v_Y  \end{matrix}\right)
\end{align*}
Straight computation yields
\begin{align*}
\Ad_{Y} \Lambda(X, u) = \Lambda(YX, \psi(Y, u)),
\end{align*}
for all $X, Y \in \SE(3) \times \R^3$ and $u^* \in \vecL^*$, so the system is equivariant.

\end{document}